\documentclass[conference]{IEEEtran}
\IEEEoverridecommandlockouts
\usepackage{graphicx} 
\usepackage{pifont}
\usepackage{amsmath}
\usepackage{amssymb}  
\usepackage{amsfonts}
\usepackage{algorithmic}
\usepackage{array}
\usepackage{multicol}

\usepackage{textcomp}
\usepackage{stfloats}
\usepackage{url}
\usepackage{verbatim}
\usepackage{graphicx}
\usepackage{caption}
\usepackage{subcaption}
\usepackage[ruled, vlined, linesnumbered]{algorithm2e}
\usepackage{cite}
\usepackage{booktabs}
\usepackage{amssymb,amsfonts,bm}
\usepackage{amsmath,tikz}
\usepackage{color}
\usepackage{mathtools}
\usepackage[hidelinks]{hyperref} 
\usepackage{rotating}
\usepackage{blkarray}
\usepackage{physics}
\usetikzlibrary{arrows}
\usepackage{makecell}

\usepackage{amsthm}
\usepackage{comment}
\usepackage{multirow}
\usepackage{geometry}
\graphicspath{{images/}}

\begin{document}

\title{\LARGE \bf
AIR-Embodied: An Efficient Active 3DGS-based Interaction and Reconstruction Framework with Embodied Large Language Model \\

\author{Zhenghao Qi, 
Shenghai Yuan, 
Jinxin Liu, 
Fen Liu, Haozhi Cao\\
Tianchen Deng,
Jianfei Yang,
and Lihua Xie,~\IEEEmembership{Fellow,~IEEE}}

\thanks{This research is supported by the National Research Foundation, Singapore, under its Medium-Sized Center for Advanced Robotics Technology Innovation (CARTIN). }
\thanks{All authors are with the School of Electrical and Electronic Engineering, Nanyang Technological University, 50 Nanyang Avenue, Singapore 639798, 
   { Email: \{shyuan, elhxie\}@ntu.edu.sg}}%
\thanks{Zhenghao Qi is also afflicted with Beihang University, Beijing, China, 100876,      
   { Email: qizh1102@163.com}}%
}

\maketitle

\begin{abstract}
Recent advancements in 3D reconstruction and neural rendering have enhanced the creation of high-quality digital assets, yet existing methods struggle to generalize across varying object shapes, textures, and occlusions. While Next Best View (NBV) planning and Learning-based approaches offer solutions, they are often limited by predefined criteria and fail to manage occlusions with human-like common sense. To address these problems, we present AIR-Embodied, a novel framework that integrates embodied AI agents with large-scale pretrained multi-modal language models to improve active 3DGS reconstruction.
AIR-Embodied utilizes a three-stage process: understanding the current reconstruction state via multi-modal prompts, planning tasks with viewpoint selection and interactive actions, and employing closed-loop reasoning to ensure accurate execution. The agent dynamically refines its actions based on discrepancies between the planned and actual outcomes.
Experimental evaluations across virtual and real-world environments demonstrate that AIR-Embodied significantly enhances reconstruction efficiency and quality, providing a robust solution to challenges in active 3D reconstruction.
\end{abstract}

\section{Introduction}


Recent advancements in 3D reconstruction and neural rendering \cite{nerf} \cite{3dgs} have greatly improved the efficiency and quality of high-quality digital assets for robot navigation, VR, AR, digital twins, gaming, and online shopping. 
While these breakthroughs offer immense potential, the ability to intelligently interact with and adapt to complex environments autonomously remains a key missing piece.

Embodied active reconstruction offers a promising approach to address the limitations of current methods. Traditional NBV planning \cite{uncertainty,fisherrf,activetutu,neu-nbv} uses predefined criteria to select optimal viewpoints from a limited set, while learning-based approaches \cite{scanrl,gennbv} attempt to improve this through reward-based policies. However, both approaches struggle with occlusions, fail to manage execution errors, and are constrained by high computational costs and poor generalization to new tasks or unseen scenarios. These challenges arise from a limited understanding of local reconstruction states and the inability to intelligently find the global optimal solutions.

The key challenge is creating an intelligent, adaptive reconstruction system that can manage real-world complexities such as occlusions and execution errors. Current methods are constrained by predefined heuristics manner, but the reasoning power of large language models presents a promising path toward more context-aware and efficient decision-making. 


\begin{figure}[t]
\centering
\includegraphics[width=\linewidth]{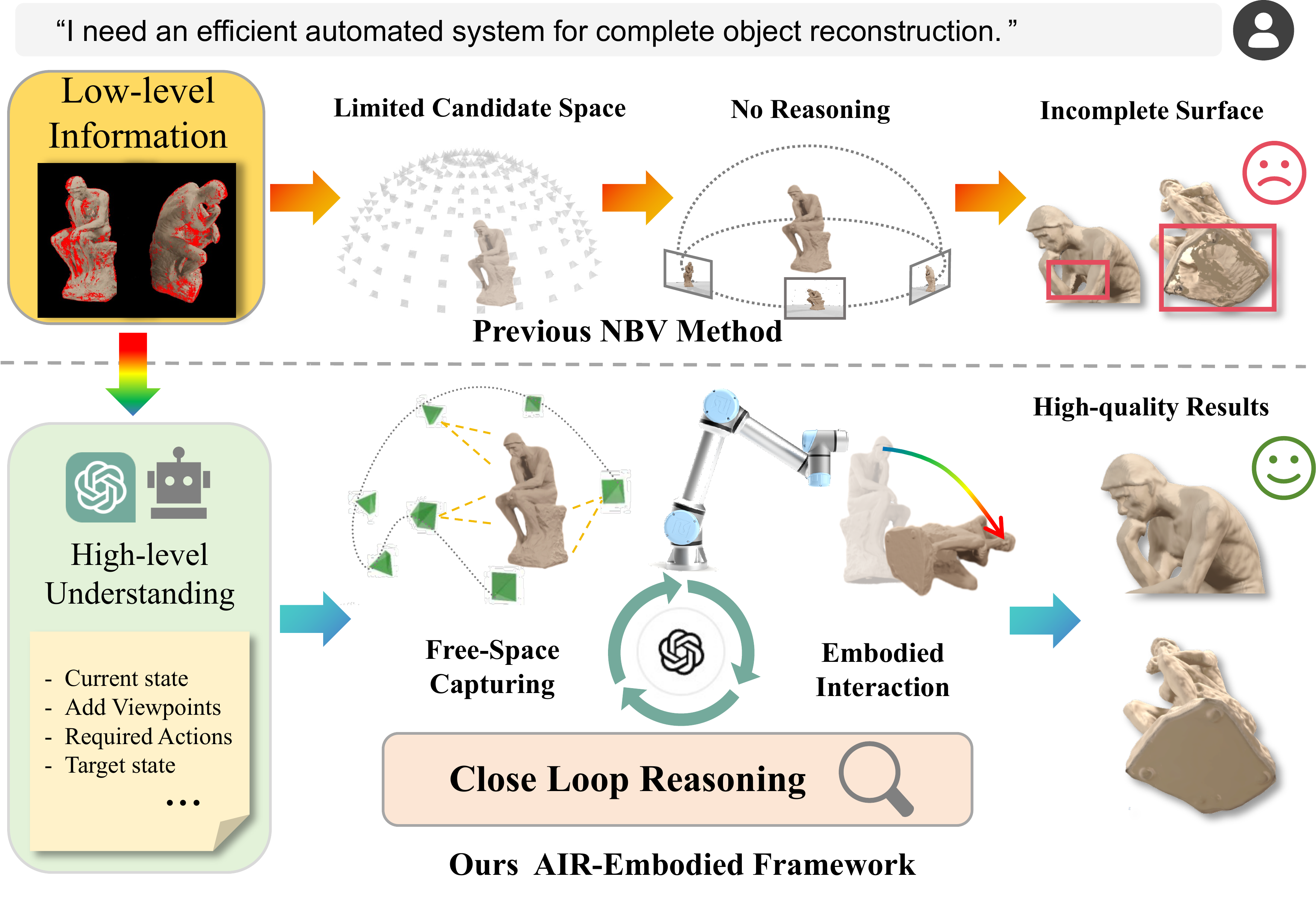} 
\caption{\textbf{Overview}. 
Previous NBV methods rely on low-level uncertainty and limited viewpoint selection. Our system uses embodied agents for high-level understanding, enabling free-space viewpoint planning and interactive manipulation. Closed-loop reasoning corrects action errors, achieving generalized, high-quality object reconstructions.
}
\label{fig:overview}
\vspace{-1em}
\end{figure}

We introduce AIR-Embodied, a framework integrating embodied AI agents with large-scale pretrained multi-modal language models (MLLM) for active 3D reconstruction, as shown in Fig. \ref{fig:overview}. The agent operates in three stages: (1) It assesses the current reconstruction state by generating multi-modal prompts from low-level pixel data and uses reasoning to identify and explain poorly reconstructed areas. (2) It plans tasks, including viewpoint selection and interactive manipulations like pushing objects to expose occluded regions. (3) The agent verifies execution results and applies closed-loop reasoning to fine-tune actions, ensuring precise reconstruction. We conducted extensive experiments on virtual datasets and real-world environments, showing that our framework greatly enhances both the efficiency and quality of active reconstruction.

Our contributions can be summarized as follows:
\begin{itemize}
  \item The paper integrates 3D Gaussian Splatting with large language models (LLMs) for viewpoint and action planning, improving the fidelity of surface representation and reconstruction quality.
  \item An optimization framework is introduced to jointly refine viewpoints and actions using a cost function, enabling efficient active task planning and execution, with a closed-loop reasoning module ensuring accuracy, quality and completeness.
  \item The system autonomously interacts with objects, using closed-loop reasoning to adapt and correct discrepancies between planned and actual actions, improving reconstruction by handling occlusions through object manipulation.
  \item Extensive experimental evaluations show that our approach outperforms the SOTA methods in terms of reconstruction quality and efficiency.
  \item We will open-source our code and methods for the benefit of the field at \url{https://github.com/QZH-00/AIR-Embodied}.
\end{itemize}

\vspace{-0.15em}
\section{Related Work}
\subsection{Active Reconstruction with Radiance Fields}
Radiance field-based 3D reconstruction has gained increasing popularity in recent years and has become a key method in fields such as virtual reality and digital assets generation. Early studies often employed neural implicit representations \cite{nerf} \cite{neus}, while more recent research has focused on explicit representations \cite{3dgs}. This approach not only enables fast rendering of high-fidelity images \cite{mip-gs} but also reconstructs high-quality geometric surfaces \cite{2dgs} \cite{pgsr}. In this paper, we adhere to this explicit representation approach.

Active 3D reconstruction was previously considered a problem of finding the Next Best View. Existing paradigms in this field can be broadly categorized into information gain-based and learning-based approaches. Information gain-based methods \cite{activermap,activetutu,neu-nbv,uncertainty,fisherrf} typically select the next view with the highest information gain based on feedback from the current reconstruction state. This can be achieved by quantifying uncertainty in the radiance field \cite{uncertainty} or using Fisher information \cite{fisherrf}. However, these methods can only capture local, low-level uncertainty and rely on manually designed rules to filter from a limited set of candidate views. Alternatively, learning-based methods, such as reinforcement learning, train view selection policies by using view coverage as the reward function \cite{scanrl} \cite{gennbv}. Neural networks have also been used to predict view planning \cite{osvp} \cite{howmany}.

However, previous paradigms are limited by their low-level understanding of the current reconstruction state.
Only NBV planning makes it difficult to fully reconstruct an object, as it struggles to handle occlusions effectively.
In contrast, our proposed active reconstruction framework leverages the reasoning and task-planning capabilities of large pre-trained models. This enables a higher-level understanding of the reconstruction state, allowing for efficient and complete object reconstruction while generalizing to previously unseen objects.

\subsection{Vision Tasks Enhanced by Embodiment}

With a physical body to control, embodied AI can significantly enhance many robotics tasks such as perception \cite{gz2} \cite{emb-seg}, tracking \cite{you}, and reconstruction\cite{agc}\cite{poke}. By active embodied robot interaction with the scene using multiple onboard sensors \cite{emb-seg}, better scene understanding can be achieved with reduced ambiguities in the virtual world. Through iterative operations, ThinkeGrasp\cite{thinkgrasp} progressively refines perception results, facilitating effective object grasping even in cluttered environments. In the field of active reconstruction, \cite{agc} utilizes robots equipped with robotic arms to grasp objects and move them in front of a depth camera to capture images from different angles. Meanwhile, object-poking method \cite{chen2023perceiving} uses implicit neural representations to discover and reconstruct unseen 3D objects, allowing robots to recognize and interact with objects in unfamiliar environments. In our framework, we additionally generate the robot's operation plan to expose the object fully.

\subsection{LLM for robotic tasks}

The integration of large language models into the field of robotics has made significant strides \cite{chatgpt-for-rob,rt2,voxposer,manipllm,look,llm+p,leo,minigpt}, particularly with the advancements in vision language models like GPT-4V \cite{yang2023dawn}. These advances have greatly improved AI's ability in understanding and reasoning \cite{chat-w-e}\cite{leo}, task planning \cite{llm+p}\cite{manipllm}, and control \cite{rt2} \cite{voxposer} in the physical world. Recent research\cite{palm} has shown that MLLM excels in logical reasoning and decision-making for complex tasks, leveraging contextual information and programming capabilities to generate effective strategies. By integrating visual data with contextual text, these models can dynamically plan and execute tasks with high success rates. However, since active reconstruction tasks are highly sensitive to perception accuracy and operational quality, posing challenges to existing methods, we have designed specialized modules to enhance MLLMs' performance in this area.

\begin{figure*}[t]
\centering
\includegraphics[width=\textwidth]{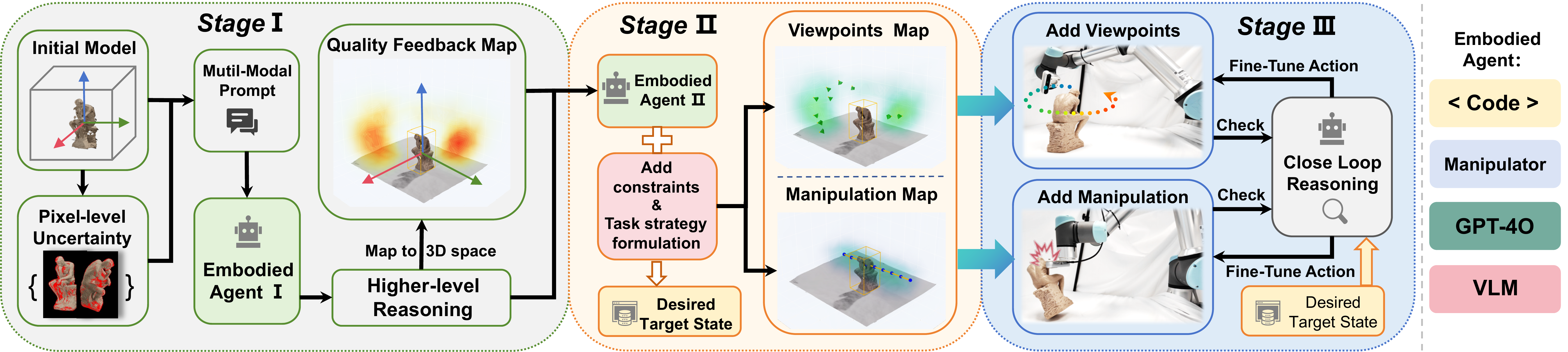} 
\caption{Overview of AIR-Embodied. In \textbf{stage I}, the agent derives high-level understanding from multi-modal low-level data and maps it to 3D space. In \textbf{stage II}, additional reasoning and constraints are added while generating plans for new viewpoints and interactive actions. In \textbf{stage III}, actions are executed, and closed-loop reasoning corrects any errors.}
\label{fig:system_overview}
\vspace{-1em} 
\end{figure*}
 
\vspace{-0.15em}
\section{Proposed Method}

\subsection{Problem Definition}
The active reconstruction problem involves reconstructing a complete 3D model of an object by selecting optimal viewpoints \( \nu = \{ v_1,v_2, \ldots, v_n\} \), \( v \in \text{SE}(3) \) and performing the required manipulations \( \tau = \{ a_1,a_2, \ldots, a_m\} \), \( a \in \text{SE}(3) \). Given an object \( \Xi \in \mathbb{R}^3 \) and its current incomplete model \( \Gamma  \in \mathbb{R}^3 \), the task is to determine the optimal viewpoints \( v \) that will most effectively fill the gaps in \( \Gamma \). The goal is to account for the uncertainty in the incomplete model within the free space and address potential occlusions, such as the bottom of the object, which may not be directly observable.

\subsection{Viewpoint and Manipulation Planning }
To address the observability problem, we first 
try to model the reconstruction uncertainty at specific positions $p_i \in \mathbb{R}^3 $.  Based on the modeled uncertainty, the system identifies the regions that require further sampling for reconstruction. After the initial assessment is done, we aim to generate a sequence of viewpoints $\nu$ and manipulations $\tau$ for an active agent to enhance task efficiency while minimizing both reconstruction error $\epsilon(\Xi, \Gamma )$ and operational cost $\zeta(\cdot)$.


Then, the active reconstruction can be formulated as an optimization problem:
\begin{equation}\label{eq: optimization}
\min_{\tau} \left\{ \epsilon(\Xi, \Gamma) + \lambda \left( \sum_{i=1}^{n} \left( \zeta_{\nu}(v_i)\right) + \sum_{j=1}^{m} \zeta_{\tau}(a_j)  \right) \right\},
\end{equation}
where $\zeta_{\nu}$ and $\zeta_{\tau}$ represent the costs of viewpoint acquisition and manipulation execution. $\lambda$ is a hyperparameter that is selected empirically to balance the action costs and reconstruction quality costs.

\subsection{Reconstruction Model Representation}
To optimize the cost function \ref{eq: optimization}, we need to define each cost item with basic representations. 
\subsubsection{Gaussian Splatting}

We employ the 3DGS representations, in which the object is explicitly represented by a set of 3D Gaussians $\{ G_i \mid i = 1, 2, \ldots, n \}$. Each Gaussian primitive is defined by a three-dimensional Gaussian function:
\begin{equation}
G_i(x | \mu_i, \Sigma_i) = e^{-\frac{1}{2} (x - \mu_i)^\top \Sigma_i^{-1} (x - \mu_i)}.
\end{equation}
The mean $\mu_i$ and covariance $\Sigma_i$ represent the position and shape of the Gaussian primitive. We follow the variant PGSR\cite{pgsr} that flattens the Gaussian sphere along the direction of the smallest scaling factor. 3DGS uses differentiable rasterization to render Gaussian primitives into 2D images, enabling parameter optimization. Given a set of Gaussians, each Gaussian primitive $G_i$ is sorted by its depth $d_i$ relative to the view plane. Then, the color of pixel $C(u,v)$ at coordinate of $(u,v)$ is obtained using alpha blending:
\begin{equation}
C(u,v) = \sum_i c_i \alpha_i \prod_{j=1}^{i-1} (1 - \alpha_j).
\end{equation}
Here, $c_i$ is the color feature vector represented by Spherical Harmonics, and $\alpha_i$ is obtained from the Gaussian weights and the Gaussian opacity parameters. Thus, the accumulated transmittance and its difference along the depth during pixel color accumulation can be expressed as:
\begin{equation}
T(i) = \alpha_i \prod_{j=1}^{i-1} (1 - \alpha_j), \quad w_i = T(i) - T(i-1)
\end{equation}
 
\subsubsection{Pixcel-level Uncertainty}
Inspired by \cite{uncertainty}, the distribution of $w_i$ along the depth $d_i$ can serve as a suitable proxy for reconstruction quality. 
For a well-reconstructed surface, pixel color is dominated by individual Gaussian primitives, reflected by concentrated weight variation $w_i$ near a complete surface, as shown in the Fig.\ref{fig:system_overview}. Therefore, we characterize the quality of the reconstruction by calculating the entropy $\xi_j$ of the depth distribution of the $j$-th Gaussian primitive weights:
\vspace{-10pt}
\begin{equation}
\xi_j = -\sum_{i=1}^{n} P(w_i,d_i) \log P(w_i,d_i).
\end{equation}
Here, $P(w_i,d_i)$ represents the presence of a significant peak in $w_i$ at depth $d_i$. The entropy reflects the concentration of peaks, indicating the uncertainty of each pixel. We set a specific threshold, and for a given viewpoint $ \mathbb{S}(R,t) \in \mathbb{R}^2$ at rotation $R$ and position $t$, the uncertainty $\Omega \in \mathbb{R}$ is defined as the ratio of pixels with values exceeding the threshold $\xi_t$ to the total number of pixels.
\begin{equation}
\Omega(\mathbb{S}(R,t)) = \frac{\sum_{1}^{N} \mathbb{I}(\xi_j > \xi_t)}{M},
\end{equation}
where $M$ is the total number of pixels in that view and $\mathbb{I}$ represents a binary function that binarizes the uncertainty. 

\subsection{AIR-Embodied Framework}
\subsubsection{High-level Reasoning}

Since pixel-level information has limited utility in view planning within 3D space, we leverage the integration and contextual reasoning abilities of large models to transform low-level information (${\mathbb{S}_p,\Omega_p}$) from broad sampling point $p \in \mathbb{R}^3$ into a high-level understanding of the current reconstruction state $\mathcal{L}_H$. To overcome the inherent ambiguity in language descriptions of free space, we voxelize the task space to map high-level understanding to 3D.

For the target object $\Xi$, the agent uses the perception module to obtain information such as its position $p_{\Xi} \in \mathbb{R}^3$ and bounding box $\Upsilon_{\Xi}$. Then, centered on $\Xi$, we construct a 3D voxel map $\gamma: \mathbb{Z}^3 \to \mathbb{R}^n$ to represent the entire task space. We perform uniform sparse sampling $\varpi$ around the object to evaluate reconstruction uncertainty:
\begin{equation}
\varpi = \{\Omega(\mathbb{S}(D_i, p_i)) \mid p_i \in \mathcal{P} \subseteq \mathbb{Z}^3\}.
\end{equation}
Here, $p_i \in \mathbb{R}^3$ is a sampled point in the voxel, and $D_i \in SO(3)$ is the rotation form of the directional vector from $p_i$ to the object's centroid. The uncertainty values are then used to initialize the voxel $\gamma = \{ \varphi_{\Omega} \}$. And $\varphi \in \mathbb{R}$ is the value function that corresponds to an uncertainty measure $\Omega$. Since small view changes have minimal impact on uncertainty sampling, we apply nearest-neighbor interpolation to fill empty voxels.
 
Next, we combine the uncertainty distribution from sampling with the rendered images to generate multi-modal prompts for the embodied agent to perform integrated reasoning. Specifically, the agent first analyzes the high-uncertainty region.
Then, the corresponding rendered images are retrieved to identify the under-reconstructed target and its causes. 
Subsequently, the data processing module augments the voxel map’s attribute values $\gamma = \{\varphi_{\Omega}, \varphi_{\text{t}}, \varphi_{\text{o}}\}$ where $t$ is the high uncertainty region and $o$ observability of this region. 
Finally, we generate the high-level understanding ${\mathcal{L}_H} = \{l_{\text{t}}, l_{\Omega}, l_{\text{o}}, l_{\text{d}}\}$ where $d$ is additional decision factor to decide on what to do. We map ${\mathcal{L}_H}$ to 3D space $\gamma \to \gamma_q$, represented as: ``Observe the object from the $\mathbb{S}(D, P)$, identify the $\varphi_\text{t}$ that are under-reconstructed". The feedback map $\gamma_q$ serves as the initial solution space to assist in subsequent view selection.

\begin{figure}[t]
\centering
\includegraphics[width=0.48\textwidth]{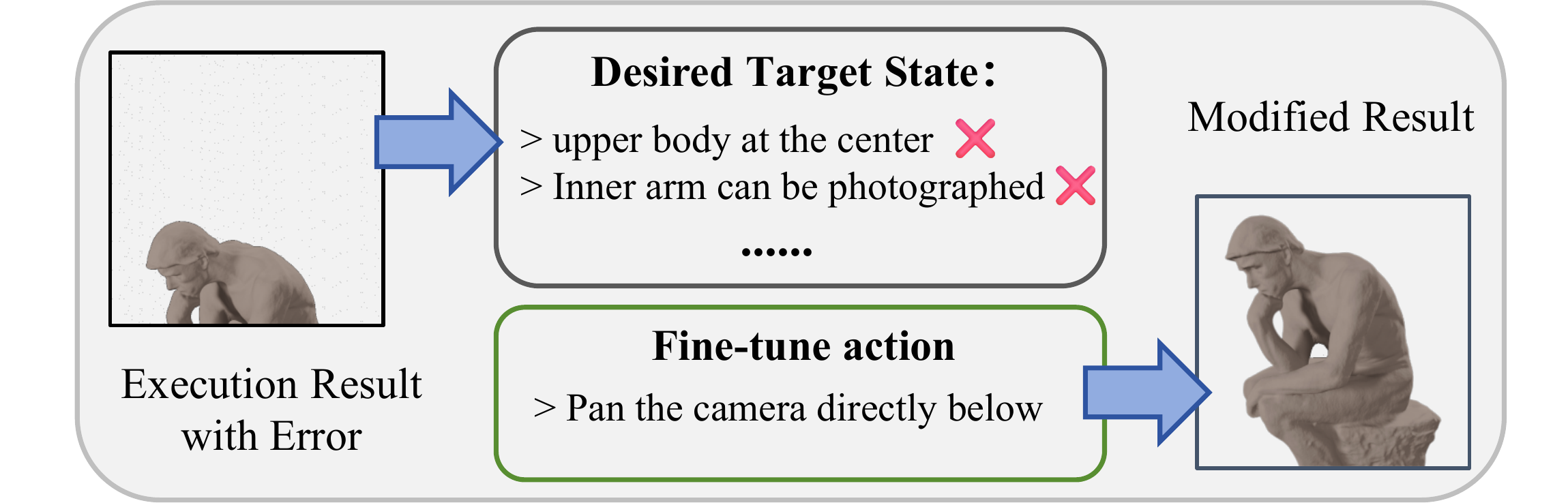} 
\vspace{-1pt}
\caption{Close Loop Reasoning. Compare the operational results with the desired target state and propose appropriate fine-tuning policy.}
\label{fig:Compensation}
\vspace{-15pt} 
\end{figure}

\subsubsection{Novel Viewpoint Synthesis}
We leverage high-level understanding $\mathcal{L}_H$ and the quality-map $\gamma_q$ to reason out the constraints for generating new views, providing action guidance for acquiring new perspectives.
The constraints inferred include task weighted constraints  $W_{\Phi}$, 
distance constraints $W_{\kappa}$, and density constraints $W_{\varkappa}$. Task constraints are derived from user-defined task requirements, determining the number and distribution of new viewpoints:
\begin{equation}
W_{\Phi}({p_i}) = \mathbb{I}({p_i} \in \varsigma ),
\end{equation}
where $\varsigma$ is generated set of option from $\mathcal{L}_H$.
The directional constraints are based on the previously identified under-reconstructed target information, determining the camera rotation $R \in \mathbb{R}^3$ for the new viewpoint, such that the camera is oriented toward the direction from the current voxel grid to the target. Setting an appropriate camera distance $\kappa_r$ at optimal threshold $r$, meaning the target occupies a suitable size in the image, the distance constraint is expressed as \eqref{e9}. The density constraint ensures that the distribution of the selected viewpoints is not overly dense, reducing redundant information and improving efficiency as \eqref{e10}. Finally, the agent scores grids in $\gamma_q$, denoted as \eqref{e11}. The grid location $p$ with the highest score is selected, forming the new viewpoint $\mathbb{S}({R}, \mathbf{t})$.
\begin{equation}
W_{\kappa}({p}) = \exp \left( -\lambda_{\kappa} (\kappa_p - \kappa_r)^2 \right),
\label{e9}
\end{equation}
\begin{equation}
W_{\varkappa}({p}) = \exp \left( -\lambda_{\varkappa} (\kappa_p - \kappa_{m})^2 \right),
\label{e10}
\end{equation}
\begin{equation}
 \varphi({p}) =  \varphi_{\Omega}({p}) \cdot W_{\Phi}({p}) \cdot W_{ \kappa}({p}) \cdot W_{\varkappa}({p}),
\label{e11}
\end{equation}
where $\kappa_p$ represents the distance value corresponding to point $p$, while $\kappa_m$ serves as the threshold for controlling the density.

\subsubsection{Manipulation Synthesis}

For the unobservable regions, typically caused by inherent occlusions, we use the high-level understanding $\mathcal{L}_H$ to plan actions that reveal these hidden areas. We input $\mathcal{L}_H$ and the designed prompts into the agent, which then reasons to generate tasks. For example, when reconstructing an object placed on a table, the contact surface between the object and the table is unobservable. In this case, the LLM breaks down the task into two subtasks: ``knock over the object" and ``supplement the viewpoint images for the object's bottom surface."

For the first subtask, the embodied agent reasons the most likely successful target location and action trajectory. Based on this, we expand the voxel map $\gamma_p$ with the attribute $\varphi_{a}$, where $\varphi_{a}(p)$ represents the score of location $p$ as a potential trajectory point for the robotic arm's end-effector.
Next, we use a greedy algorithm to search for a set of discrete trajectory points from all the high-scoring points. Then, we use MoveIt to connect these discrete points, generating a smooth action trajectory.

After completing the action, since the object's state has changed, the system re-invokes the perception module and, through reasoning, maps the values of the original voxel map $\gamma_p$ to the new voxel map $\gamma_{\text{new}}$ in the current state. At this point, the system modifies the relevant attributes in the voxel grid based on the object's current state. Subsequently, new viewpoints are selected again.

\subsubsection{Close Loop Reasoning}

The reconstruction task imposes stringent requirements on the camera's pose during capture. However, due to perception errors and control inaccuracies, open-loop guidance often results in deviations from the desired pose, as shown in Fig. \ref{fig:Compensation}. To address this, we designed a closed-loop reasoning and verification module.
After each action $\tau$ is completed, our closed-loop reasoning agent compares the captured results, current scene state $\mathcal{L}^{S}_i$, and desired state $\mathcal{L}^{S}_i$ to evaluate task completion. The agent computes the discrepancy $\Delta \mathcal{L}^{S} = \mathcal{L}^{S}_i - \mathcal{L}^{S}_{desired}$ to assess the accuracy of the execution. 
If the action results do not meet the expected requirements, the system will perform fine-tuning and corrections based on the discrepancies between the actual operation results and the target state, ensuring precise task completion.

\begin{table*}[t]
\centering
\caption{OminiObject3D Simulation Experiment, Best results are in \textbf{bold}, second best are \underline{underlined}.}
\vspace{-4pt}
\label{table1}
\centering
\renewcommand{\arraystretch}{1.15}
\begin{tabular}{cccccccccc} 
\hline\hline
                           & Methods            & ~ ~PSNR↑~ ~     & ~ ~SSIM↑~ ~    & ~ ~LPIPS↓~ ~   & ~ ~Acc↓~ ~      & ~ ~Comp↓~ ~     & Chamfer↓        & ~ F-score↑~~    & ACR↑          \\ 
\hline
\multirow{2}{*}{Heuristic} & Random \cite{choy20163d}             & 25.567          & 0.744          & 0.276          & 0.0214          & 0.0182          & 0.0198          & 0.4953          & 2.28\%           \\
                           & Uniform \cite{schmid2020efficient}           & 28.563          & 0.757          & 0.262          & 0.0094          & 0.0117          & 0.0105          & 0.6578          & 4.14\%           \\ 
\hline
\multirow{3}{*}{NBV}  & Uncertainty \cite{uncertainty} & 28.481          & 0.757         & 0.263         & 0.0095          & 0.0121          & 0.0108          & 0.6302          & 4.02\%           \\
                           & FisherRF \cite{fisherrf}           & 28.687          & 0.757          &  \underline{0.260}          & 0.0087          & 0.0106          & 0.0096          & 0.7648          & 4.32\%           \\
                           & Ours w/o Manip      &\underline{29.113}           & \underline{0.773}           & 0.269          & \underline{0.0079}            & \underline{0.0099  }        & \underline{0.0089 }         & \underline{0.8377}          & \underline{4.66}\%           \\ 
\hline
Embodied                   & Ours         & \textbf{30.846} & \textbf{0.790} & \textbf{0.241} & \textbf{0.0059} & \textbf{0.0057} & \textbf{0.0058} & \textbf{0.9237} & \textbf{5.08\%}  \\
\hline
\end{tabular}
\end{table*}

\begin{table*}[t]
\centering
\caption{Real-World Experiment, Best results are in \textbf{bold}, second best are \underline{underlined}.}
\vspace{-4pt}
\label{table2}
\centering
\setlength{\tabcolsep}{5pt} 
\renewcommand{\arraystretch}{1.15}
\begin{tabular}{c|cccc|cccc|cccc} 
\hline\hline
\multirow{3}{*}{Methods} & \multicolumn{4}{c|}{Simple}                                           & \multicolumn{4}{c|}{Medium}                                           & \multicolumn{4}{c}{Complex}                                            \\ 
\cline{2-13}
                         & \multicolumn{8}{c|}{20 Views}                                                                                                                 & \multicolumn{4}{c}{30 Views}                                           \\ 
\cline{2-13}
                         & Acc↓            & ~Comp↓          & Chamfer↓        & ACR↑            & Acc↓            & ~Comp↓          & Chamfer↓        & ACR↑            & Acc↓            & ~Comp↓          & Chamfer↓        & ACR↑             \\ 
\hline
Random \cite{choy20163d}              & 0.0121          & 0.0137          & 0.0199          & 2.01\%          & 0.0274          & 0.0244          & 0.0259          & 2.52\%          & 0.1021         & 0.1171          & 0.1091          & 1.91\%           \\           
Uniform \cite{schmid2020efficient}                  & 0.0071          & 0.0087          & 0.0079          & 2.96\%          & 0.0195          & 0.0197          & 0.0196          & 4.01\%          & 0.0498          & 0.0341          & 0.0419          & 3.56\%           \\
Uncertainty  \cite{uncertainty}               & 0.0071          & 0.0083          & 0.0076          & 2.96\%          & 0.0199          & 0.0198          & 0.0198          & 3.99\%          & 0.0501          & 0.0354          & 0.0427          & 3.43\%           \\
FisherRF \cite{fisherrf}                & \underline{0.0069}          & 0.0083          & 0.0075          & 2.98\%          & \underline{0.0164}          & 0.0178          & 0.0176          & 4.10\%          & 0.0475          & 0.0311          & 0.0393          & 3.89\%           \\
Ours w/o Loop                & 0.0071          & \underline{0.0064}          & \underline{0.0067}          & \underline{3.01\%}          & 0.0173          & \underline{0.0145}          & \underline{0.0159}          & \underline{4.25}\%          & \underline{0.0473}          & \underline{0.0271}          & \underline{0.0372}          & \underline{4.19\%}           \\
Ours                     & \textbf{0.0061} & \textbf{0.0058} & \textbf{0.0059} & \textbf{3.04\%} & \textbf{0.0103} & \textbf{0.0097} & \textbf{0.0100} & \textbf{4.97\%} & \textbf{0.0341} & \textbf{0.0257} & \textbf{0.0299} & \textbf{4.98\%}  \\
\hline\hline
\end{tabular}
\vspace{-8pt}
\end{table*}

\begin{figure}[t]
\centering
\includegraphics[width=0.48\textwidth]{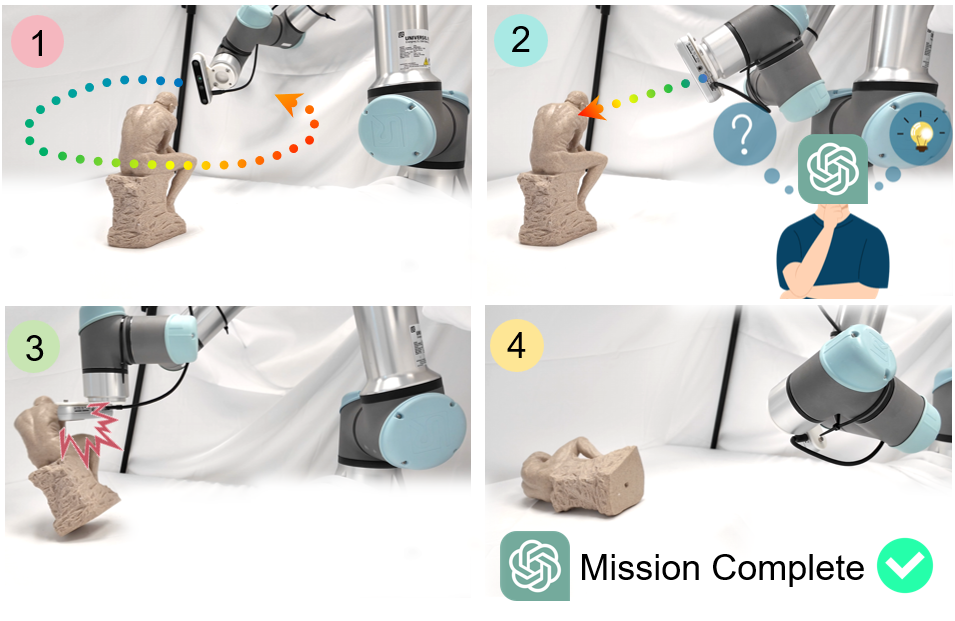} 
\vspace{-8pt}
\caption{ AIR-Embodied: active reasoning while scanning.}
\label{fig:realword_demo}
\vspace{-1em} 
\end{figure}

\vspace{-0.15em}
\section{Experiment}

We evaluated our proposed framework using both virtual and real-world experiments across a diverse range of objects, as shown in Fig. \ref{fig:realword_demo}. The experimental results demonstrate the effectiveness and generalization of the proposed method.

\subsection{Datasets and Metrics}
\noindent \textbf{Simulation Experiment}: We utilized the OmniObject3D dataset for our simulation experiments. OmniObject3D offers high-fidelity models of 190 objects scanned from the real world. We used 50 of these objects to evaluate the zero-shot generalization capability of our method.\\
\textbf{Real-world Experiment}: For real-world experiments, we selected three categories of items based on structural and texture complexity: everyday product packaging, 3D-printed sculptures, and intricate artifacts. These categories were chosen to demonstrate our framework's adaptability and potential for real-world applications.\\
\textbf{Metrics}: We evaluated our framework and other baseline methods using three sets of metrics. We used PSNR, SSIM, and LPIPS to assess rendering quality. Additionally, we used Accuracy, Completeness, Chamfer, and F-score to evaluate geometric quality. To assess the efficiency of viewpoint selection, we employed the Average Contribution Rate, which measures the average contribution of each newly selected viewpoint to the improvement of model quality. It is worth noting that the image test set in the simulation experiments includes 120 images randomly sampled from the spherical space around the objects, while the ground truth models in real-world experiments are derived from CAD drawings.
\subsection{Baselines Selection}

To comprehensively compare the advantages of our proposed framework, we selected the following methods as baselines. Heuristic methods include random sampling \cite{choy20163d} and uniform sampling \cite{schmid2020efficient}, where images are captured along a fixed preset trajectory. Uncertainty-based \cite{uncertainty} and FisherRF \cite{fisherrf} are information gain-based methods that iteratively select the optimal viewpoints from a predefined set of candidates on a spherical surface. 
Additionally, we conducted ablation studies, including using our framework for viewpoint planning only, and using our framework without the closed-loop reasoning module.

\subsection{Implementation Details}

\noindent \textbf{Reconstruction Setup:} 
For all comparison methods that initially used NeRF, we uniformly replaced their approaches with 3DGS \cite{pgsr} for a fair comparison. The experimental setup is consistent, with all methods initialized using the same four images and the same initial point cloud, and limited the total number of viewpoints to 20. For the heuristic methods and our framework, new viewpoints were selected after an initial 10,000 iterations. For the other methods, viewpoint selection was conducted according to the settings described in their original papers \cite{uncertainty} \cite{fisherrf}. 

\noindent \textbf{LLM and APIs Setup:} We used GPT-4O and followed the prompt structure of VoxPoser \cite{voxposer} to generate code and recursively call the LLM API. We predefined basic perception and data Processing APIs, including functions for obtaining the 3D bounding box of objects, the centroid of objects, the 2D mask of objects. as well as reading and modifying Voxels.

\noindent \textbf{Hardware:} Our experiments were conducted on a computer equipped with an i7-13700 CPU and an NVIDIA RTX 4070 Super GPU. Both the simulation and real-world experiments were performed using a UR5e robotic arm and a RealSense D455 camera.

\subsection{Simulator Experiment}
We conducted our simulation experiments in Isaac SIM. The URDF models of the OminiObject were exported from Blender. We used the robotic arm to collect a spherical candidate set required for the baseline methods. Our experimental results are reported in Tab \ref{table1}. As shown in \ref{table1}, our framework achieved state-of-the-art results across a wide range of object categories. 
Compared to the baselines, we obtained better rendering results, with significant improvements in geometric accuracy and completeness. Meanwhile, our average contribution rate also performed the best. These improvements are thanks to our more flexible viewpoint selection and the complete exposure of objects through interactive operations. Notably, even when using our method solely for viewpoint planning, we still achieved second-best results in most metrics, which demonstrates the efficiency gains brought by our free-space viewpoint planning approach.

\begin{figure}[t]
\centering
\includegraphics[width=0.55\textwidth]{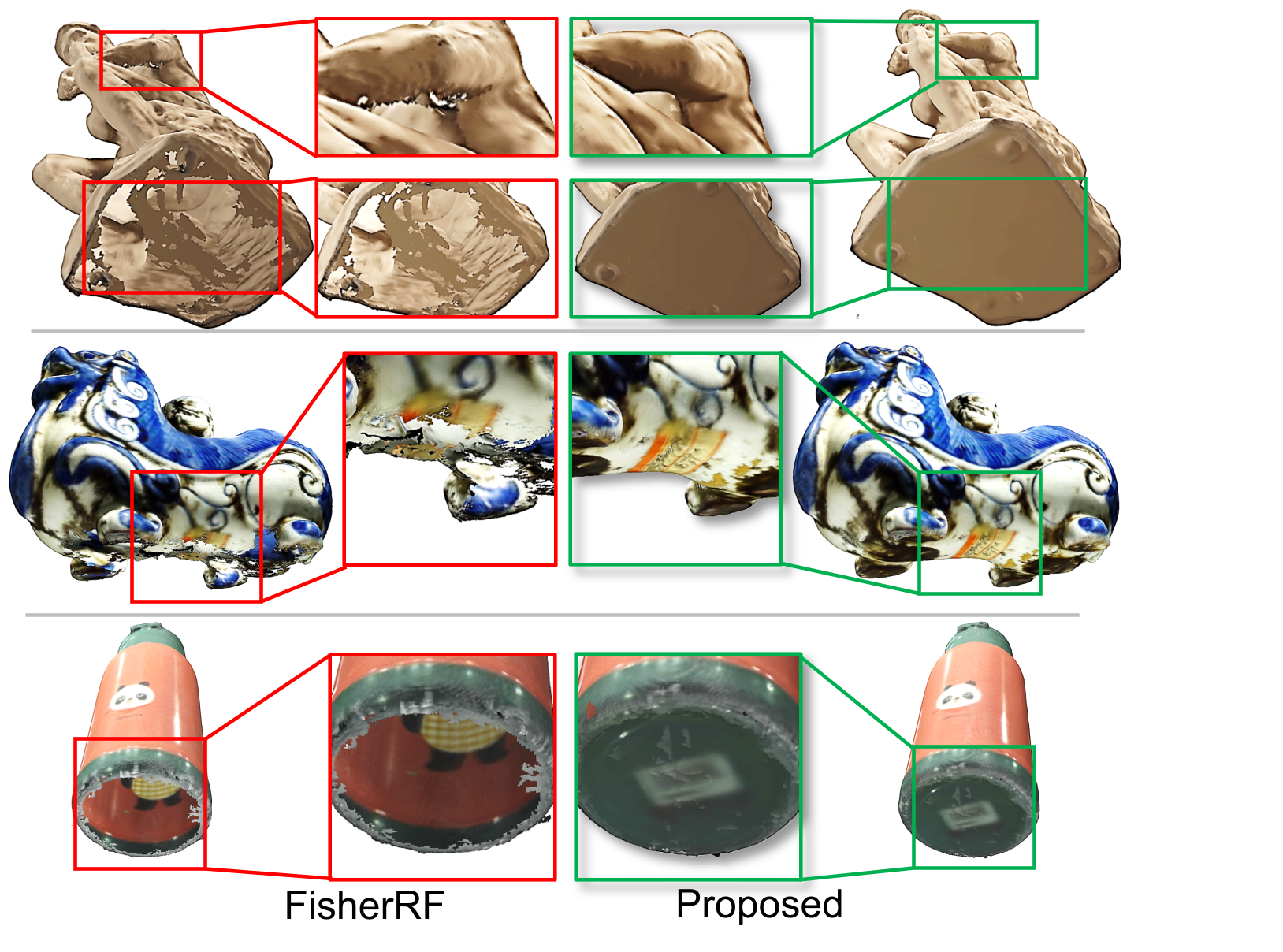}
\vspace{-15pt} 
\caption{Qualitative Comparison. The proposed method scans better than the current SOTA.}
\label{comparsionz}
\vspace{-1em} 
\end{figure}

Qualitative results, as shown in the Fig.\ref{comparsionz}, indicate that our framework reveals more object information and more complete geometric structures compared to methods that use only viewpoint planning, which is crucial for many application tasks.

In our experiments, we found that vanilla LLMs tend to provide programmatic and repetitive answers for viewpoint selection tasks. Therefore, we discussed the flexibility of our framework and conducted 15 different experiments to measure the repetitiveness of LLM responses. 
The results, shown in Table \ref{table3}, indicate that neither GPT-4O-only nor GPT-4-based methods can offer targeted solutions for this task. In contrast, our method can intelligently select viewpoints through guidance based on reconstruction quality feedback. Additionally, our results highlight the significant importance of the closed-loop reasoning module in our framework for tasks that are sensitive to pose.

\begin{table}
\centering
\caption{Suitability Experiment}
\label{table3}
\centering
\renewcommand{\arraystretch}{1.15}
\begin{tabular}{c|cccc} 
\hline\hline
              & Promots & APIs & Repetition Rate↓ & ACR↑            \\ 
\hline
GPT-4O only   & \checkmark       & \checkmark    & 73.3\%           & 3.21\%           \\
Voxposer      & \checkmark       & \checkmark    & 66.7\%           & 3.27\%           \\
Ours w/o loop & \checkmark       & \checkmark    & 13.3\%           & 4.24\%           \\
Ours          & \checkmark       & \checkmark    & \textbf{13.3\%}  & \textbf{4.89\%}  \\
\hline\hline
\end{tabular}
\vspace{-10pt}
\end{table}

\subsection{Real-World Experiment}

To assess the Sim2Real capability of our method, we conducted real-world experiments, with the results reported in Table \ref{table2}. We performed three sets of experiments ranging from simple to complex, all captured by the UR5e robotic arm after viewpoint planning. Despite perception and control errors in the real-world environment posing challenges to open-loop methods, our closed-loop reasoning module allows our approach to maintain state-of-the-art performance and generalize across objects of varying complexity. Furthermore, the operations generated by our approach ensure that our completeness and overall performance consistently remain superior, demonstrating its potential for practical application.

\section{Conclusion}
In this work, we introduced AIR-Embodied, a novel framework that integrates large-scale multi-modal language models with embodied AI agents for active 3D reconstruction. Through extensive experiments in both virtual and real-world environments, we demonstrated significant improvements in reconstruction efficiency and quality, as shown in Fig. \ref{comparsionz}. By combining viewpoint planning, interactive manipulations, and closed-loop reasoning, our approach effectively addresses occlusions and execution errors, pushing the boundaries of autonomous reconstruction systems.

\bibliographystyle{IEEEtran}
\bibliography{mybib}


\end{document}